\documentclass[acmtog]{acmart}
\acmSubmissionID{2234}

\usepackage{booktabs}
\usepackage{pifont}
\usepackage[ruled]{algorithm2e}
\usepackage{multirow}
\usepackage{threeparttable}

\SetAlFnt{\small}
\SetAlCapFnt{\small}
\SetAlCapNameFnt{\small}
\SetAlCapHSkip{0pt}

\acmJournal{TOG}
\begin{document}
\title{\textbf{S}cene and  \textbf{H}uman in \textbf{O}ne \textbf{W}orld: Reconstruction in a Feedforward Pass}

\author{Boao Shi}
\email{boaoshi@seas.upenn.edu}
\orcid{0009-0008-4386-1421}
\affiliation{%
  \institution{University of Pennsylvania}
  \city{Philadelphia}
  \state{Pennsylvania}
  \country{USA}
}

\author{Qiao Feng}
\email{fengqiao@seas.upenn.edu}
\orcid{0000-0003-0625-651X}
\affiliation{%
  \institution{University of Pennsylvania}
  \city{Philadelphia}
  \state{Pennsylvania}
  \country{USA}
}

\author{Yiming Huang}
\email{ymhuang9@seas.upenn.edu}
\orcid{0009-0004-4001-0630}
\affiliation{%
  \institution{University of Pennsylvania}
  \city{Philadelphia}
  \state{Pennsylvania}
  \country{USA}
}

\author{Lingjie Liu}
\email{lingjie.liu@seas.upenn.edu}
\orcid{0000-0003-4301-1474}
\affiliation{%
  \institution{University of Pennsylvania}
  \city{Philadelphia}
  \state{Pennsylvania}
  \country{USA}
}

\newcommand{\YH}[1]{{\color{blue}YH:#1}}
\newcommand{\LJ}[1]{{\color{red}LJ:#1}}

\begin{abstract}

Reconstructing humans in dynamic scenes from moving monocular cameras remains challenging due to scale ambiguity, human–scene misalignment, and occlusion interference. Rather than treating human mesh recovery and scene reconstruction as separate tasks, we believe that accurate human–scene reconstruction requires the two tasks to mutually inform each other: parametric human models offer semantic structure and metric-scale priors, while scene geometry provides spatial context for human localization and alignment. Built on this insight, we introduce SHOW, a mask-promptable human mesh recovery framework that couples feed-forward 3D scene reconstruction with Human Mesh Recovery in a unified metric space. SHOW injects human semantics and scale priors from parametric human models into normalized point-map prediction, enabling metric-scale scene reconstruction from inherently scale-ambiguous monocular input. In turn, the recovered scene geometry constrains human mesh estimation, encouraging spatially consistent human placement and improved human–scene alignment. To handle complex multi-person and cluttered scenes, SHOW further incorporates a promptable masking mechanism that enables flexible target-human selection while suppressing background distractions and occlusion interference. Through joint training, the model learns both human-aware geometric features and geometry-constrained human features, producing aligned metric-scale reconstructions from monocular human-centric videos. Extensive experiments demonstrate that SHOW improves metric-scale consistency, human–scene alignment, and reconstruction accuracy under challenging camera motion, occlusion, and cluttered backgrounds. Project page: \href{https://bowieshi.github.io/SHOW-project-page/}{\textcolor{cyan}{https://bowieshi.github.io/SHOW-project-page/}}

\end{abstract}



\begin{teaserfigure}
\centering
\includegraphics[width=\textwidth]{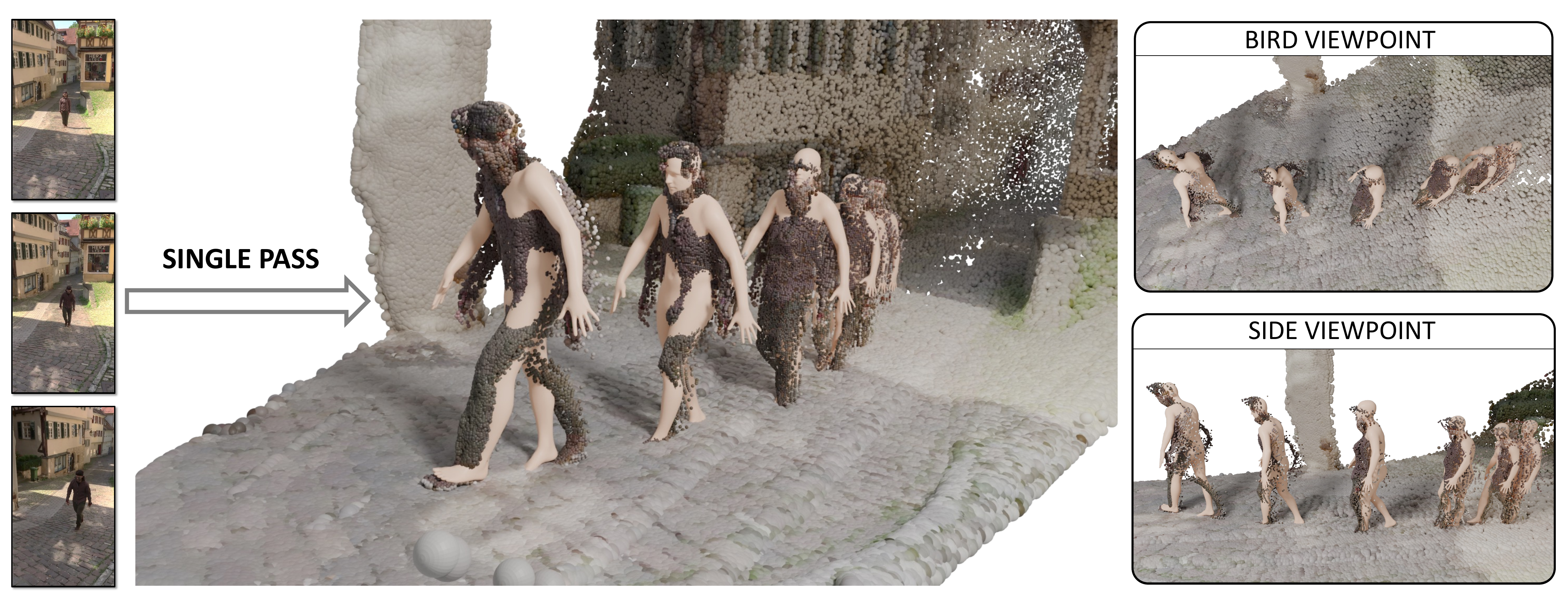}
\caption{Our method can produce better human and scene alignment under monocular human-centric video input. Our method will predict human parametric model and scene point map in a feedforward manner. 
}
\label{fig:teaser}
\end{teaserfigure}

\maketitle

\section{Introduction}
In the real world, humans move, act, and interact within dynamic environments. Their behaviors are shaped by the surrounding scene, while their actions can also alter the environment itself. Humans and scenes therefore coexist in a shared physical world. Accurately modeling their spatial and physical relationships is essential for understanding grounded human behavior for various downstream applications including AR/VR, gaming, and robotics \cite{wang2023scene,yi2024tesmo, wang2025physhsi, xie2022chore, xie2023vistracker, xie2026cari4d, xie2024InterTrack}.

Despite this close coupling, many existing reconstruction pipelines still treat humans and scenes as separate targets, often estimating humans and scenes independently before aligning them in a common coordinate system \cite{liu2026joint, xue2024hsr}.  In practice, such decoupled pipelines often rely on fragile localization cues, such as heuristic body anchors, detector-based estimates, or post-hoc alignment, and use scene geometry only as auxiliary context rather than as an active constraint for human reconstruction. As a result, the reconstructed human and scene may appear reasonable separately, but remain inconsistent in scale, position, or physical interaction when placed together. This issue is especially severe for monocular human-centric videos, where depth and scale ambiguity make it difficult to determine whether the recovered body is correctly localized with respect to the surrounding scene.

Recent feed-forward methods have begun to address this limitation by reconstructing humans and scenes within a unified pipeline. For example, Human3R~\cite{chen2025human3r} leverages the strong 3D priors of visual geometry models to directly recover humans, scenes, and camera trajectories in a single forward pass. UniSH~\cite{li2026unishunifyingscenehuman} further targets metric-scale human-scene reconstruction by combining scene and human reconstruction branches with scale and localization fusion. While these methods demonstrate the effectiveness of applying geometry foundation models for human-scene reconstruction, their coupling between the human and scene components remains limited. In particular, the geometry model is largely used as a general-purpose 3D predictor, while the human model is used mainly to provide body estimates or metric-scale cues for post-alignment. As a result, human priors are not explicitly injected into the geometry model to learn human-aware scene features, and scene geometry is not deeply fused back into the human reconstruction process to constrain body pose, scale, and global placement.

In contrast, we argue that human-scene reconstruction should be formulated as an explicitly coupled problem, where humans and scenes mutually inform each other throughout the reconstruction process. Our key insight is that since humans and scenes coexist in the same physical world, they should be reconstructed jointly. Motivated by this, we propose SHOW, a mask-promptable feed-forward framework for human-scene aligned reconstruction from monocular human-centric videos. Rather than treating scene reconstruction and human mesh recovery as separate prediction tasks, SHOW couples them through two complementary mechanisms. 

First, SHOW makes scene reconstruction human-aware by adapting a pretrained visual geometry foundation model for human-centric reconstruction. Specifically, we fine-tune the geometry backbone with an auxiliary DensePose head, encouraging it to learn dense correspondences between image pixels and the human body surface and to produce geometric features better aligned with human shape and pose. A human-mask prompting mechanism further enables identifying the target person and focusing feature extraction on the relevant human region. In this way, human semantic cues and body-scale priors are incorporated into scene point-map prediction, helping resolve monocular scale ambiguity and recover scene geometry in a metric space consistent with humans.

Second, SHOW makes human recovery scene-aware by fusing visual geometry features with HMR features for SMPL-X estimation and global localization. These scene-aware features provide spatial context for placing the reconstructed body within the surrounding environment, encouraging the estimated human mesh to be metrically consistent with the recovered scene.

Through joint training, SHOW predicts dense scene geometry, scene scale, SMPL-X body parameters, and global human placement in a shared metric coordinate system, as illustrated in Fig.~\ref{fig:teaser}. The bidirectional coupling between human and scene enables human-aware geometry reconstruction and geometry-constrained human recovery within a unified feed-forward model.

Our contributions are summarized as follows:
\begin{itemize}
    \item We introduce SHOW, a unified feed-forward framework for monocular human-scene reconstruction that jointly estimates dense scene geometry and SMPL-X human meshes in a shared metric coordinate system, reducing scale ambiguity and cascading errors from multi-stage alignment pipelines.

    \item We inject human semantic and metric-scale priors into scene reconstruction, while using scene geometry to constrain human mesh recovery, enabling joint learning of human-aware geometric features and geometry-constrained human features for better human-scene alignment.

    \item We propose a mask-as-prompt mechanism that uses human masks as dense 2D prompts for visual geometry foundation models, enabling target-aware feature extraction, improved human surface localization, and stronger robustness to clutter and multi-person interference.

    \item We provide a comprehensive evaluation protocol for human scene alignment, with extensive benchmarking of existing feed-forward methods to support future research.
\end{itemize}

\begin{figure*}[t]
    \centering
    \makebox[\textwidth][c]{%
        \includegraphics[width=0.95\textwidth, trim=0 0 0 0, clip]{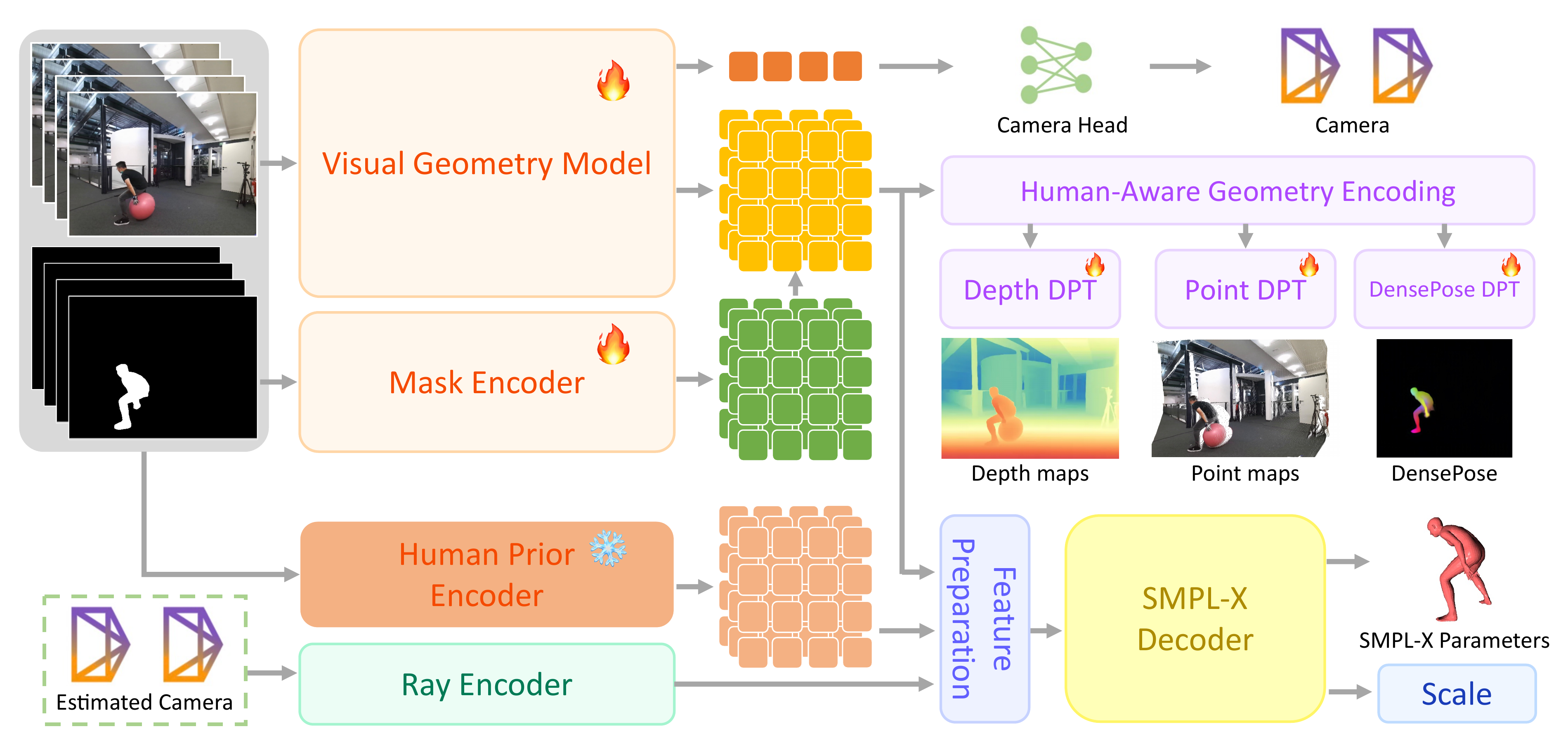}
    }
    \caption{Given a monocular video, SHOW reconstructs visible humans and the surrounding scene in a shared 3D world through a single feed-forward pipeline. We first adapt VGGT with mask prompting and an auxiliary DensePose head to extract human-aware geometry features and predict normalized scene point maps. A geometry-aware SMPL-X decoder then fuses human features with scene context to estimate body pose, shape, global translation, and a scale factor that maps the normalized point map to the metric scale of each reconstructed human. Through joint training, SHOW aligns human scale and placement with the surrounding 3D scene, producing coherent human-scene reconstructions.}
    \label{fig:transformer_architecture_comparison}
\end{figure*}

\section{Related Work}

\subsection{World-Grounded Human Mesh Recovery}

Traditional Human Mesh Recovery (HMR) methods primarily estimate human pose and shape in the camera coordinate frame \cite{hmrKanazawa17,humanMotionKZFM19,kocabas2019vibe,patel2024camerahmr,goel2023humans,yang2026sam3dbody}. While effective for local body articulation, camera-space HMR does not explicitly resolve the metric scale, depth, and global placement ambiguities in monocular 3D reconstruction. World-grounded HMR methods address this by lifting human motion into a gravity-aware world frame \cite{WHAM,yuan2022glamr,TRAM}, typically using relative motion prediction or global camera rotation estimation to recover temporally coherent human trajectories in world coordinates. However, these methods remain largely human-centric. They aim to recover coherent global human trajectories, but do not explicitly align the reconstructed human with surrounding scene geometry. As a result, they may achieve temporal consistency and gravity alignment while still suffering from scene-level metric misalignment, incorrect placement, or long-term drift, especially when global motion is estimated autoregressively \cite{shen2024gvhmr}.

\subsection{Foundation Models for 3D/4D Reconstruction}
Recent visual geometry foundation models, such as CUT3R \cite{cut3r} and VGGT \cite{wang2025vggt}, have introduced strong feed-forward priors for 3D and 4D reconstruction. By jointly predicting geometric signals such as camera parameters, depth, point maps, and scene structure, these models provide a unified representation for recovering scenes from visual observations. Building on these advances, recent works have adapted such geometry foundation models to human-centric reconstruction. Human3R \cite{chen2025human3r} extends the online 4D reconstruction model CUT3R with parameter-efficient visual prompt tuning, enabling SMPL-X body parameters to be directly decoded from reconstruction latents. Similarly, JOSH3R \cite{liu2026joint} augments the 3R reconstruction paradigm with a dedicated human tracking head to improve human-aware 4D reconstruction. Despite their effectiveness, existing methods often depend on coarse human localization cues, such as head detection or other sparse heuristics, to associate humans with the reconstructed scene. These cues can be unreliable under occlusion, truncation, crowded scenes, or unusual viewpoints, leading to inaccurate human-scene alignment. This motivates the need for more robust and alignment-aware conditioning mechanisms. In our work, we address this limitation through a mask-based prompting strategy that provides stronger spatial guidance for grounding human reconstruction within the surrounding 3D/4D scene.

\subsection{Human-Scene Alignment}

Accurately modeling the relationship between humans and their surrounding scene is essential for physically grounded human reconstruction \cite{cao2026hsimul3r, wang2025physhsi, videomimic}. To achieve human-scene alignment, early works often rely on contact-based optimization or conditioning, either to populate reconstructed scenes with plausible human bodies \cite{hassan_samp_2021, Hassan:CVPR:2021} or to synthesize scenes conditioned on human poses \cite{li2024physics, nie2021pose2room}. More recent human-scene reconstruction methods move toward feedforward frameworks for jointly recovering humans and their environment. UniSH \cite{li2026unishunifyingscenehuman} integrates scene reconstruction with HMR priors to recover coherent human meshes together with scene geometry, while Human3R \cite{chen2025human3r} enables online 4D reconstruction by estimating global multi-person meshes and dense scene geometry in a single pass. Despite this progress, accurate metric alignment between reconstructed humans and the surrounding environment remains challenging. We argue that effective human-scene alignment requires more than jointly predicting humans and scenes in the same output space: the two reconstruction tasks must actively inform each other during learning. To this end, our method injects human semantics and body-scale priors into scene geometry estimation, while using scene-aware geometric features to constrain human mesh recovery and global localization.

\section{Method}
Given a monocular video $\mathcal{I}=\{I_t\}_{t=1}^{T}$, our goal is to reconstruct humans and scenes together in a shared 3D world in a single feed-forward pass. 
For each frame $I_t$, we predict a normalized partial 3D point map 
$\bar{\mathbf{P}}_t=\{\bar{\mathbf{p}}_{t,u}\}_{u\in\Omega}$ and reconstruct all visible humans 
$\mathcal{H}_t=\{H_{t,i}\}_{i=1}^{N_t}$ in the same coordinate frame. 
Here $\Omega$ denotes the set of image pixels, and $\bar{\mathbf{p}}_{t,u}\in\mathbb{R}^3$ is the normalized 3D point corresponding to pixel $u$. 
Each human instance is represented as 
$H_{t,i}=\{\theta_{t,i}, \beta_{t,i}, \tau_{t,i}, s_{t,i}\}$, where $\theta_{t,i}$, $\beta_{t,i}$, and $\tau_{t,i}$ denote pose, shape, and global translation, respectively. 
The scalar $s_{t,i}$ maps the normalized point map to the scale of the reconstructed human body, i.e., $\mathbf{P}_{t,i}=s_{t,i}\bar{\mathbf{P}}_t$. 
This scale factor allows the reconstructed humans to be geometrically aligned with the surrounding scene.

Figure~\ref{fig:transformer_architecture_comparison} provides an overview of our pipeline. 
First, we adapt VGGT~\cite{wang2025vggt} for human-aware geometry encoding by incorporating a mask encoder and an auxiliary DensePose head (Sec.~\ref{sec:human-prior-vggt}).
Second, we train a geometry-aware SMPL-X decoder that combines human features~\cite{wang2025prompthmr} with scene context to recover body parameters in the reconstructed 3D space (Sec.~\ref{sec:smpl-decoder}). 
Finally, we jointly fine-tune the human and scene reconstruction modules so that each recovered body scale and placement is aligned with the surrounding 3D scene (Sec.~\ref{sec:joint-training}).

\subsection{Human-Aware Geometry Encoding}
\label{sec:human-prior-vggt}

VGGT~\cite{wang2025vggt} provides strong feed-forward geometry features for camera, depth, and point-map prediction. 
However, its latent geometry features are optimized for generic geometry reconstruction, and the predicted depth and point maps are not necessarily aligned with the real-world scale or the scale of human body models.
This scale mismatch makes the raw VGGT features less suitable for directly guiding human body scale and placement.
The key idea is to adapt the geometry encoding so that its latent features retain human-related cues useful for downstream SMPL-X decoding. 
We incorporate a lightweight mask encoder and supervise the latent features with an auxiliary DensePose head. 
This encourages the geometry features to recognize those regions as human bodies, providing stronger cues for estimating human scale and placement in the reconstructed 3D space.

\textbf{Mask Encoder.}
For each frame $I_t$, we first obtain a human foreground mask $M_t$ using an off-the-shelf segmentation model~\cite{ravi2024sam2, carion2025sam3segmentconcepts}. 
The mask is encoded by a lightweight mask encoder:
\begin{equation}
    F_t^{m} = E_{msk}(M_t).
\end{equation}
In parallel, the VGGT encoder extracts latent geometry features from the input frame:
\begin{equation}
    F_t^{g} = E_{geo}(I_t),
\end{equation}
where $E_{geo}$ denotes the VGGT image encoder. 
We incorporate the mask feature into the latent geometry feature through an adaptation module:
\begin{equation}
    \tilde{F}_t^{g} = A(F_t^{g}, F_t^{m}),
\end{equation}
where $\tilde{F}_t^{g}$ is the resulting human-aware geometry feature. 
The adapted feature $\tilde{F}_t^{g}$ is then used for depth and point-map prediction, and later passed to the SMPL-X decoder to provide scene-level cues for body scale and placement.

\begin{figure}[t]
    \centering
    \includegraphics[width=\linewidth]{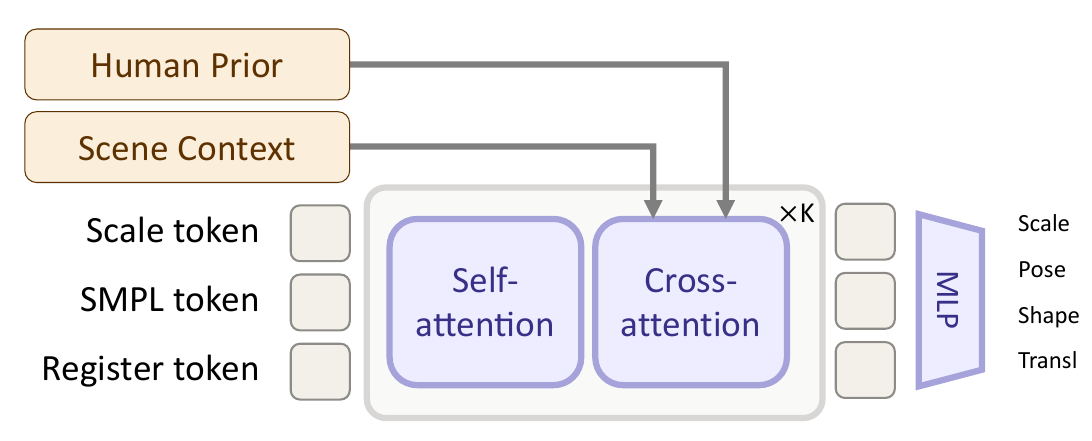}
    \caption{SMPL-X decoder architecture. Scale, SMPL, and register tokens will go through K-times self-attention and cross-attention with positional-encoding-added geometry and human patch features. The updated tokens are fed into an MLP to predict scale, pose, shape, and translation.}
    \Description{decoder}
    \label{fig:decoder}
\end{figure}

\textbf{DensePose Auxiliary Supervision.}
Adding a mask encoder alone does not guarantee that the geometry features become human-aware, since the network may still treat the mask as a generic foreground cue. 
To explicitly encourage the latent geometry feature to encode human body information, we supervise $\tilde{F}_t^{g}$ with an auxiliary DensePose prediction task:
\begin{equation}
    \hat{\mathbf{Y}}_t = H_{\mathrm{dp}}(\tilde{F}_t^{g}),
\end{equation}
where $\hat{\mathbf{Y}}_t$ denotes the predicted DensePose map. 
Given the DensePose target $\mathbf{Y}_t$, we use a foreground-weighted regression loss:
\begin{equation}
    \mathcal{L}_{\mathrm{dp}}
    =
    \frac{1}{|\Omega|}
    \sum_{u\in\Omega}
    w_{t,u}
    \left\|
        \hat{\mathbf{Y}}_{t,u}
        -
        \mathbf{Y}_{t,u}
    \right\|_2^2 ,
\end{equation}
where $w_{t,u}$ upweights human pixels to prevent the loss from being dominated by the background. We set $w_{t,u}=1/r_t$ for human pixels and $w_{t,u}=1/(1-r_t)$ otherwise, where $r_t$ is the foreground pixel ratio.
This auxiliary supervision forces the shared latent feature to preserve information that is predictive of human body regions, instead of merely separating foreground from background. 
The DensePose prediction is used only during training, while the resulting feature $\tilde{F}_t^{g}$ is passed to the SMPL-X decoder at inference time.

We train the human-aware geometry encoding with both the original VGGT geometry losses and the auxiliary DensePose loss. 
The DensePose loss injects human-specific supervision, while the geometry losses prevent degradation in camera, depth, and point-map prediction:
\begin{equation}
    \mathcal{L}_{\mathrm{geo}}
    =
    \mathcal{L}_{\mathrm{camera}}
    +
    \mathcal{L}_{\mathrm{depth}}
    +
    \mathcal{L}_{\mathrm{pmap}}
    +
    \lambda_{\mathrm{dp}}\mathcal{L}_{\mathrm{dp}},
\end{equation}
where $\lambda_{\mathrm{dp}}$ is the weight for the DensePose auxiliary loss, and the remaining geometry losses follow VGGT~\cite{wang2025vggt}.
The resulting features retain VGGT's scene-level geometry capability while becoming more useful for downstream SMPL-X decoding.

\subsection{Geometry-Aware SMPL-X Decoding}
\label{sec:smpl-decoder}
After obtaining the human-aware geometry features from Sec.~\ref{sec:human-prior-vggt}, we decode SMPL-X parameters with explicit scene context. 
As shown in Fig.~\ref{fig:decoder}, the decoder takes two complementary feature streams: human-centric features that provide body pose and shape cues, and adapted geometry features that provide scene context for body scale and placement.

\textbf{Feature Preparation.}
For each human instance $i$ in frame $I_t$, we extract human feature tokens from PromptHMR~\cite{wang2025prompthmr}:
\begin{equation}
    F_{t,i}^{h} = E_{\mathrm{phmr}}(I_t, m_{t,i}),
\end{equation}
where $m_{t,i}$ is the instance prompt. 
We pair these tokens with the adapted geometry feature $\tilde{F}_{t}^{g}$ from Sec.~\ref{sec:human-prior-vggt}. 
To provide camera-aware spatial information, we add ray-based positional encodings computed from the predicted camera intrinsics:
\begin{equation}
    \bar{F}_{t,i}^{h} = F_{t,i}^{h} + \mathrm{RayEnc}(\hat{\mathbf{K}}_t), 
    \qquad
\end{equation}
The resulting human tokens $\bar{F}_{t,i}^{h}$ and geometry tokens $\tilde{F}_{t}^{g}$ are used as inputs to the geometry-aware SMPL-X decoder.

\textbf{Geometry-Aware Token Decoder.}
We use a lightweight Transformer decoder to regress SMPL-X parameters from the prepared human and geometry tokens. 
For each human instance, we initialize three query tokens:
\begin{equation}
    Q_{t,i}^{0}
    =
    \{q_{t,i}^{\mathrm{smpl}}, q_{t,i}^{\mathrm{scale}}, q_{t,i}^{\mathrm{reg}}\}.
\end{equation}
The SMPL token $q_{t,i}^{\mathrm{smpl}}$ is responsible for body pose, shape and global translation, the scale token $q_{t,i}^{\mathrm{scale}}$ estimates point map scale parameter, and the register token $q_{t,i}^{\mathrm{reg}}$ facilitates information exchange.

Each decoder layer first performs self-attention among the three query tokens, followed by cross-attention to the human and geometry features:
\begin{equation}
    Q_{t,i}^{l+1}
    = \mathrm{CrossAttn}
    \left(
        \mathrm{SelfAttn}(Q_{t,i}^{l});
        \bar{F}_{t,i}^{h},
        \tilde{F}_{t}^{g}
    \right).
\end{equation}
In practice, we use an asymmetric cross-attention design. The SMPL and register tokens attend to both human features $\bar{F}_{t,i}^{h}$ and geometry features $\bar{F}_{t}^{g}$, while the scale token attends only to geometry features.
This encourages pose and shape estimation to use human priors, while grounding scale in the reconstructed scene. 

After $K$ decoder layers, lightweight MLP heads regress the final parameters:
\begin{equation}
\begin{split}
    (\hat{\theta}_{t,i}, \hat{\beta}_{t,i}, \hat{\tau}_{t,i})
    &= \mathrm{Head}_{\mathrm{smpl}}(q_{t,i}^{\mathrm{smpl},K}), \\
    (\hat{s}_{t,i})
    &= \mathrm{Head}_{\mathrm{geo}}
    (q_{t,i}^{\mathrm{scale},K}).
\end{split}
\end{equation}
The SMPL head predicts body pose, shape and translation, while the geometry head predicts scale for aligning reconstructed normalized 3D space with the parametric body.

\textbf{Decoder Training.}
In the second training stage, we freeze the human-aware geometry encoder and train only the SMPL-X decoder with standard human reconstruction losses:
\begin{equation}
    \mathcal{L}_{\mathrm{hmr}}
    =
    \lambda_{2d}\mathcal{L}_{2d}
    +
    \lambda_{3d}\mathcal{L}_{3d}
    +
    \lambda_{\mathrm{smpl}}\mathcal{L}_{\mathrm{smpl}}
    +
    \lambda_{v}\mathcal{L}_{v}
    +
    \lambda_{\tau}\mathcal{L}_{\tau}
    +
    \lambda_{s}\mathcal{L}_{s}.
\end{equation}
This stage teaches the decoder to recover body pose, shape, translation, and scale from the combined human and geometry features, while the next stage further enforces their alignment with the reconstructed scene. Additional details are provided in the supplementary material.

\subsection{Joint Training for Human-Scene Alignment}
\label{sec:joint-training}

After the first two stages, the geometry encoder provides human-aware scene features and the SMPL-X decoder predicts human bodies from both human and scene cues. 
However, training these two components separately does not explicitly enforce consistency between the reconstructed body and the predicted 3D point map. 
We therefore perform a final joint fine-tuning stage to align the human reconstruction with the surrounding scene geometry.

\textbf{Scale and Placement Consistency.}
For each human instance, the decoder predicts a scale factor $\hat{s}_{t,i}$ and a global translation $\hat{\tau}_{t,i}$. 
Here $\hat{s}_{t,i}$ rescales the normalized point map to be consistent with the reconstructed human body, while $\hat{\tau}_{t,i}$ places the body in the reconstructed 3D space. 
We supervise the scale factor in log space:
\begin{equation}
    \mathcal{L}_{\mathrm{scale}}
    =
    \left\|
    \log \hat{s}_{t,i}
    -
    \log s_{t,i}^{*}
    \right\|_2^2,
\end{equation}
where $s_{t,i}^{*}$ denotes the ground-truth scale factor.

For video inputs, the same person should have a stable scale across frames. 
We therefore regularize the predicted scale toward its sequence-level average:
\begin{equation}
    \bar{s}_{i}
    =
    \frac{1}{|\mathcal{T}_i|}
    \sum_{t\in\mathcal{T}_i}
    \hat{s}_{t,i},
\end{equation}
\begin{equation}
    \mathcal{L}_{\mathrm{temp}}
    =
    \sum_{t\in\mathcal{T}_i}
    \left\|
    \log \hat{s}_{t,i}
    -
    \log \bar{s}_{i}
    \right\|_2^2,
\end{equation}
where $\mathcal{T}_i$ denotes the frames in which instance $i$ appears.

\textbf{SMPL-Guided Point-Map Alignment.}
In addition to scale consistency, we use the reconstructed human surface to regularize the point-map geometry around human regions. 
Since the point map is predicted in a normalized scale, we first rescale the human-region point map using the predicted scale factor:
\begin{equation}
    \hat{\mathbf{P}}_{t,i}^{h}
    =
    \hat{s}_{t,i}\,\hat{\bar{\mathbf{P}}}_{t}[m_{t,i}],
\end{equation}
where $\hat{\bar{\mathbf{P}}}_{t}[m_{t,i}]$ denotes the normalized point-map region inside the human mask. 
We render the predicted SMPL-X mesh to obtain its visible surface points $\hat{V}_{t,i}^{\mathrm{vis}}$. 
Since the visibility operation is non-differentiable, we detach these visible surface points and use them as geometric anchors for the point map. 
The alignment loss is defined as
\begin{equation}
    \mathcal{L}_{\mathrm{align}}
    =
    d_{\mathrm{chamfer}}
    \left(
        \hat{\mathbf{P}}_{t,i}^{h},
        \mathrm{sg}(\hat{V}_{t,i}^{\mathrm{vis}})
    \right),
\end{equation}
where $\mathrm{sg}(\cdot)$ denotes stop-gradient. 
This loss pulls the rescaled point-map geometry toward the detached visible body surface, encouraging the human-region point map to be compatible with the reconstructed SMPL-X body.

\textbf{Joint Fine-tuning Objective.}
The final training objective combines the geometry loss, the SMPL-X reconstruction loss, and the human-scene alignment losses:
\begin{equation}
    \mathcal{L}_{\mathrm{total}}
    =
    \mathcal{L}_{\mathrm{geo}}
    +
    \mathcal{L}_{\mathrm{hmr}}
    +
    \lambda_{\mathrm{scale}}\mathcal{L}_{\mathrm{scale}}
    +
    \lambda_{\mathrm{temp}}\mathcal{L}_{\mathrm{temp}}
    +
    \lambda_{\mathrm{align}}\mathcal{L}_{\mathrm{align}}.
\end{equation}
This stage jointly fine-tunes the human and scene reconstruction modules, encouraging the normalized point map, the predicted scale factor, and the reconstructed body surface to be consistent in the shared 3D space.

\section{Experiments}

\subsection{Implementation Details}

During the stage one training (Sec.~\ref{sec:human-prior-vggt}) , we train the visual geometry backbone using 4 NVIDIA B200 GPUs for 100 epochs. 
In the subsequent SMPL-X decoder training stage (Sec.~\ref{sec:smpl-decoder}), we train the decoder using 2 NVIDIA B200 GPUs for another 100 epochs. In the final end-to-end training stage (Sec.~\ref{sec:joint-training}), we train the model using 4 NVIDIA B200 GPUs for another 200 epoch.

\subsection{Datasets and Metrics}

\paragraph{Datasets.}
We train our model on \textbf{BEDLAM2.0}, a large-scale synthetic dataset that provides accurate SMPL-X parameters, depth annotations, and camera trajectories for human mesh recovery in complex environments. To ensure reliable geometric supervision, we discard sequences with inaccurate HDRI-based depth rendering, resulting in 4,250 valid sequences. During training, we downsample videos to 6 FPS and use 5-frame clips to increase temporal diversity.

\paragraph{Metrics.}
For human mesh recovery, we follow standard evaluation protocols in both camera and world coordinate spaces. In camera space, we evaluate local pose and shape accuracy on 3DPW and EMDB (subset 1) using Mean Per Joint Position Error (MPJPE), Procrustes-aligned MPJPE (PA-MPJPE), and Per Vertex Error (PVE), all measured in millimeters. In world space, we evaluate long-term motion on EMDB (subset 2) and RICH, where each sequence is split into 100-frame segments. We report World MPJPE (W-MPJPE), World-aligned MPJPE (WA-MPJPE) and Root Translation Error (RTE, in \%) to assess global trajectory accuracy.

To evaluate human–scene consistency, we propose two new metrics: \textbf{Human–Scene Chamfer (HS-CF)} and \textbf{Human–Scene Variance (HS-V)}. HS-CF measures the normalized chamfer distance between body model points and human point map (scene point map under human mask). HS-V compares the spatial variance of body model points and human point map along horizontal and vertical axes in camera frame. Together, these metrics assess whether the reconstructed human and scene are properly aligned in a unified space.

\begin{table}[t]
\centering
\scriptsize
\setlength{\tabcolsep}{4.3pt}
\renewcommand{\arraystretch}{0.9}
\begin{tabular}{ll ccc ccc}
\toprule
Cat. & Method &
\multicolumn{3}{c}{3DPW} &
\multicolumn{3}{c}{EMDB-1} \\
\cmidrule(lr){3-5} \cmidrule(lr){6-8}
& & PA & MPJ & PVE & PA & MPJ & PVE \\
\midrule

w/o sc.
& CLIFF~\cite{li2022cliff}       & 43.0 & 69.0 & 81.2 & 68.3 & 103.3 & 123.7 \\
& HMR2.0a~\cite{goel2023humans}  & 44.4 & 69.8 & 82.2 & 61.5 & 97.8  & 120.0 \\
& TokenHMR~\cite{dwivedi_cvpr2024_tokenhmr} & 44.3 & 71.0 & 84.6 & 55.6 & 91.7  & 109.4 \\
& CameraHMR~\cite{patel2024camerahmr} & 38.5 & 62.1 & 72.9 & 43.7 & 73.0 & 85.4 \\
& NLF~\cite{sarandi2024nlf}      & \underline{37.3} & \underline{60.3} & \underline{71.4} & \underline{41.2} & \textbf{69.6} & \textbf{82.4} \\
& PromptHMR~\cite{wang2025prompthmr} & \textbf{36.6} & \textbf{58.7} & \textbf{69.4} & \textbf{41.0} & \underline{71.7} & \underline{84.5} \\

\midrule

w/ sc.
& Human3R~\cite{chen2025human3r} & 44.1 & 71.2 & 84.9 & 48.5 & \textbf{73.9} & \textbf{86.0} \\
& UniSH~\cite{li2026unishunifyingscenehuman} & 48.8 & 75.6 & 88.8 & 86.6 & 112.9 & 140.0 \\
& \textbf{Ours} & \textbf{41.0} & \textbf{67.7} & \textbf{78.7} & \textbf{45.8} & 75.7 & 88.8 \\

\bottomrule
\end{tabular}
\caption{\textbf{Local mesh reconstruction results.}
PA: PA-MPJPE (mm). Lower is better for all metrics.}
\label{tab:local}
\vspace{-10pt}
\end{table}

\begin{table}[t]
\centering
\footnotesize
\setlength{\tabcolsep}{1pt}
\renewcommand{\arraystretch}{0.95}
\begin{tabular}{lccccc ccc}
\toprule
Method & Opt. Free & Scene &
\multicolumn{3}{c}{EMDB-2} &
\multicolumn{3}{c}{RICH} \\
\cmidrule(lr){4-6} \cmidrule(lr){7-9}
& & &
WA$\downarrow$ & W$\downarrow$ & RTE$\downarrow$ &
WA$\downarrow$ & W$\downarrow$ & RTE$\downarrow$ \\
\midrule

SLAHMR~\cite{ye2023slahmr} & \ding{55} & \ding{55} & 326.9 & 776.1 & 10.2 & 132.2 & 237.1 & 6.4 \\
TRAM~\cite{TRAM} & \ding{55} & \ding{55} & 76.4 & 222.4 & 1.4 & 127.8 & 238.0 & 6.0 \\
JOSH~\cite{liu2026joint} & \ding{55} & \ding{51} & \textbf{68.9} & \textbf{174.7} & \textbf{1.3} & \textbf{89.0} & \textbf{132.5} & \textbf{3.0} \\
\midrule

TRACE~\cite{TRACE} & \ding{51} & \ding{55} & 429.0 & 1702.3 & 17.7 & 238.1 & 925.4 & 101.4 \\
WHAM~\cite{WHAM} & \ding{51} & \ding{55} & 135.6 & 334.8 & 6.0 & 108.4 & 190.1 & 4.5 \\
GVHMR~\cite{shen2024gvhmr} & \ding{51} & \ding{55} & \textbf{111.0} & \textbf{276.5} & \textbf{2.0} & \textbf{78.8} & \textbf{126.3} & \textbf{2.4} \\
\midrule

JOSH3R~\cite{liu2026joint} & \ding{51} & \ding{51} & 220.0 & 661.7 & 13.1 & -- & -- & -- \\
Human3R~\cite{chen2025human3r} & \ding{51} & \ding{51} & 112.2 & 267.9 & 2.2 & 110.0 & 184.9 & 3.3 \\
UniSH~\cite{li2026unishunifyingscenehuman} & \ding{51} & \ding{51} & 118.5 & 270.1 & 5.8 & 118.1 & 183.2 & 4.8 \\
\textbf{Ours} & \ding{51} & \ding{51} & \textbf{109.1} & \textbf{262.3} & \textbf{2.1} & \textbf{107.3} & \textbf{172.7} & \textbf{2.2} \\
\bottomrule
\end{tabular}
\caption{\textbf{Global human motion estimation on EMDB-2 and RICH.}
Opt. indicates feed-forward (optimization-free).
Scene indicates joint 3D scene reconstruction.
WA: World-aligned MPJPE, W: MPJPE, RTE: Root translation error (\%).}
\label{tab:global_hmr}
\end{table}

\begin{table}[t]
\centering
\footnotesize
\setlength{\tabcolsep}{3pt}
\renewcommand{\arraystretch}{1}

\begin{threeparttable}
\begin{tabular}{lcccccccc}
\toprule
Method &
\multicolumn{4}{c}{3DPW} &
\multicolumn{4}{c}{EMDB-1} \\
\cmidrule(lr){2-5} \cmidrule(lr){6-9}
& HS-V$_5$ & HS-V$_{10}$ & HS-CF$_5$ & HS-CF$_{10}$
& HS-V$_5$ & HS-V$_{10}$ & HS-CF$_5$ & HS-CF$_{10}$ \\
\midrule
Human3R$^{1}$ 
& 0.652 & 0.640 & 1.30 & 1.32
& 3.66 & 2.91 & 1.13 & 1.31 \\
\midrule
UniSH$^{2}$   
& \textbf{0.028} & \underline{0.028} & \underline{7.65} & \underline{7.52} 
& \underline{0.033} & \underline{0.037} & \underline{7.10} & \underline{6.99} \\

Ours
& \underline{0.030} & \textbf{0.026} & \textbf{5.69} & \textbf{5.44} 
& \textbf{0.013} & \textbf{0.017} & \textbf{5.88} & \textbf{5.79} \\
\bottomrule
\end{tabular}

\begin{tablenotes}
\footnotesize
\item[1] scene point branch trained with additional metric scale point map data.
\item[2] scale branch trained with additional self collected real world video data.

\end{tablenotes}

\end{threeparttable}

\caption{Comparison of human-scene consistency metrics on 3DPW and EMDB1. Lower is better for all metrics.}
\label{tab:human_scene_metrics}
\end{table}

\textbf{Human-Scene-Chamfer (HS-CF).} HS-CF is defined as:
\begin{equation}
\text{HS-CF} = \frac{\sum_{i=1}^{N} \min_{p \in \mathcal{S}} \| h_i - p \|_2 }{ \sum_{i=1}^{N} | z_i^\text{human} - \bar{z}^\text{human} | },
\end{equation}
where $\mathcal{H} = \{h_i\}_{i=1}^N$ denotes the set of body model visible points; $\mathcal{S}$ denotes the filtered\footnote{To reduce boundary outliers and errors from imperfect human masks, we keep only points within the 5\%–95\% or 10\%–90\% percentile range, which becomes $HS\text{-}CF_5$, $HS\text{-}V_5$ and $HS\text{-}CF_{10}$, $HS\text{-}V_{10}$.} human point map; $z_i^\text{human}$ is the $z$-coordinate of the human visible point $h_i$ in camera frame, and $\bar{z}^\text{human}$ is the average $z$-coordinate of all visible human points.

\textbf{Human-Scene-Variance (HS-V).}
We first compute the variance of the body model visible points along the horizontal ($x$) and vertical ($y$) axes in camera frame. Similarly, we compute the variance of the human point map along the same axes. The HS-V is then defined as the average $\ell_1$ distance between these variances:

\begin{equation}
\mathrm{HS\text{-}V} = \frac{1}{2} \left( 
\left| \mathrm{Var}_x(\mathcal{V}_h) - \mathrm{Var}_x(\mathcal{P}_h) \right| +
\left| \mathrm{Var}_y(\mathcal{V}_h) - \mathrm{Var}_y(\mathcal{P}_h) \right|
\right),
\end{equation}

\begin{table}[t]
\centering
\footnotesize
\resizebox{\columnwidth}{!}{
\begin{tabular}{lcccc}
\toprule
Method &
\multicolumn{4}{c}{3DPW} \\
\cmidrule(lr){2-5}
& HS-V$_5\downarrow$ & HS-V$_{10}\downarrow$ & HS-CF$_5\downarrow$ & HS-CF$_{10}\downarrow$ \\
\midrule
w/o pretrain alignment
& 0.708 & 0.660 & 38.51 & 38.39 \\
w/o joint modeling
& 0.312 & 0.295 & 17.32 & 17.08 \\
w/o mask prompting
& 0.033 & 0.035 & 14.12 & 14.10 \\
w/o ray encoder
& 0.734 & 0.667 & 11.23 & 11.06 \\
full
& \textbf{0.028} & \underline{0.030} & \underline{9.44} & \underline{9.38} \\

full + e2e tuning
& \underline{0.029} & \textbf{0.027} & \textbf{6.21} & \textbf{5.99} \\
\bottomrule
\end{tabular}
}
\caption{Comparison of human–scene consistency metrics on 3DPW (lower is better). Models are evaluated on subsampled 6 FPS data and trained for 150 epochs in the ablation setting; end-to-end fine-tuning is performed for an additional 100 epochs.}
\label{tab:ablation}
\end{table}
\subsection{Comparison}
\paragraph{Local human mesh recovery.} 
We compare our method with state-of-the-art human mesh recovery approaches, as shown in Table~\ref{tab:local}. Our method achieves the best performance on most metrics among joint human–scene feed-forward approaches. We attribute this improvement to our mask-prompting mechanism, which provides more precise and robust human localization. Human3R relies on head detection for human estimation; when the head is occluded or in uncommon pose, it may fail to produce predictions, leading to significant performance degradation. Our method leverages full-body masks as prompts, allowing reliable pose estimation even when only part of the body is visible. UniSH uses bounding box prompts, which can be ambiguous in scenarios involving close human interactions, where multiple individuals overlap. In such cases, bounding boxes are insufficient to disambiguate the target person, whereas our mask-based representation enables more accurate and instance-specific estimation.

\paragraph{Global human mesh recovery.}
We compare our method with prior approaches for global human mesh recovery, as shown in Table~\ref{tab:global_hmr}. Our method outperforms existing joint human–scene feed-forward approaches and achieves performance on par with state-of-the-art methods. Notably, on EMDB-2, which features moving cameras, our approach demonstrates clear advantages. By modeling human and scene in a unified space, we achieve more accurate global motion estimation, even surpassing methods such as GVHMR that rely on an off-the-shelf scale estimator. On the RICH dataset, however, our performance is slightly lower. We attribute this to its static-camera setup, where visual geometry models typically benefit more from multi-view observation, which is not satisfied in the RICH dataset.

\begin{figure*}[!t]
    \centering
\includegraphics[width=0.9\linewidth]{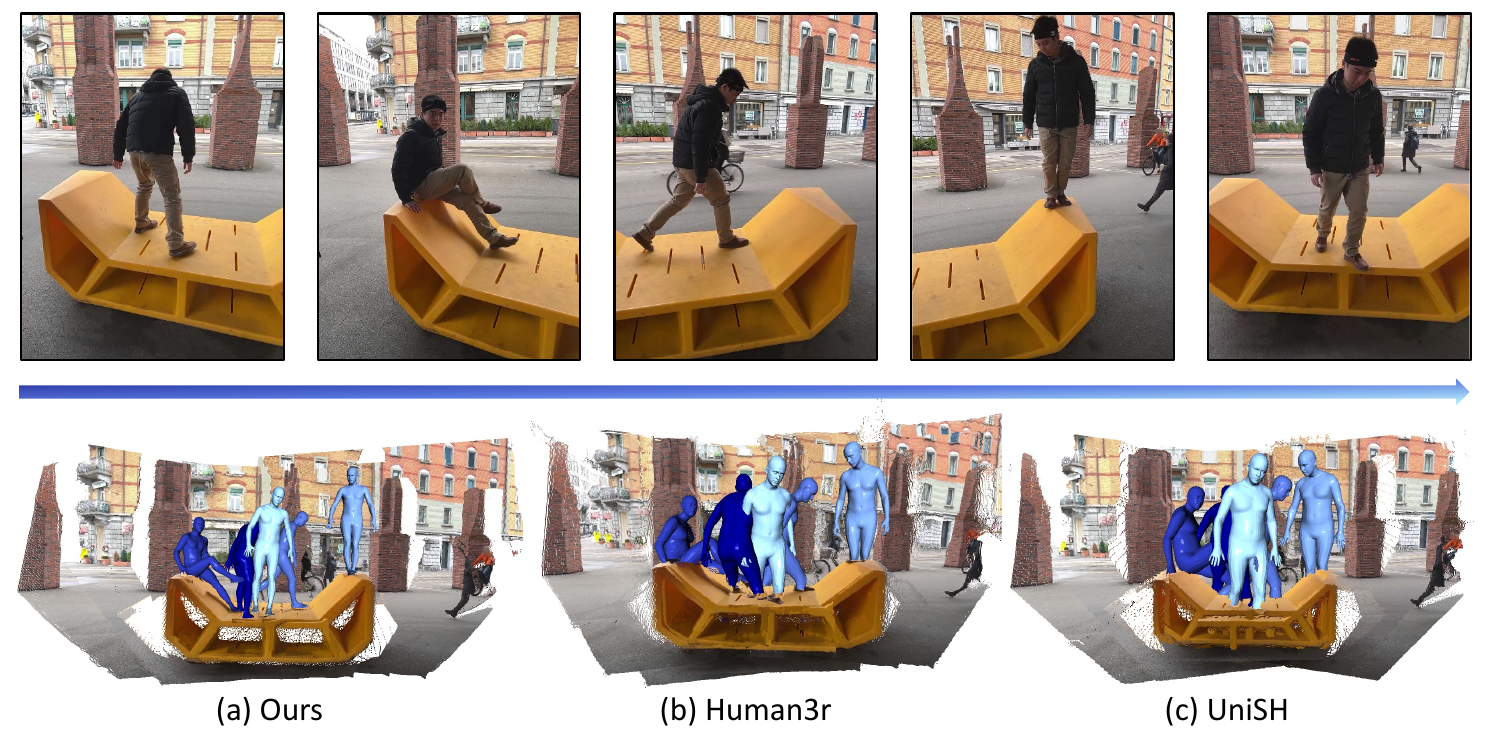}
\caption{
Qualitative Comparison of Human-Scene Reconstruction in video. Given a monocular human-centric video (top), we compare our method against Human3r and UniSH. Our framework demonstrates superior compatibility between the reconstructed SMPL-X body models and the scene point map. While baseline methods exhibit significant human-scene mismatch-such as the human body penetrating the bench geometry—our approach achieves physically plausible alignment and interaction.
}
  \Description{human_scene}
    \label{fig:human_scene}
\end{figure*}

\begin{figure*}[!t]
    \centering
\includegraphics[width=0.95\linewidth]{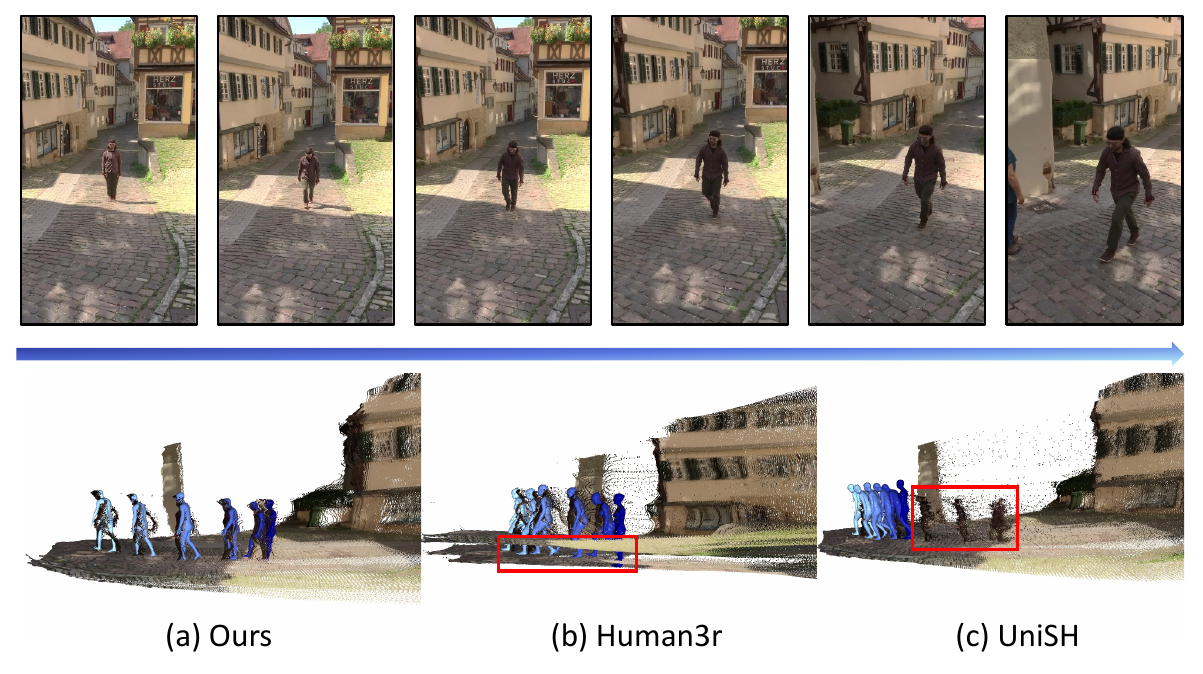}
\caption{
Qualitative comparison of human point map and human body model alignment in video. Given a monocular human-centric video (top), we compare our method with Human3r and UniSH. Our method achieves accurate alignment with the human point map. In contrast, Human3r exhibits noticeable ground penetration of the human body model, while UniSH fails to properly align the body model with the human point map over large regions.
}
  \Description{human_point}
    \label{fig:human_point}
\end{figure*}

\begin{figure*}[!t]
    \centering
\includegraphics[width=\linewidth]{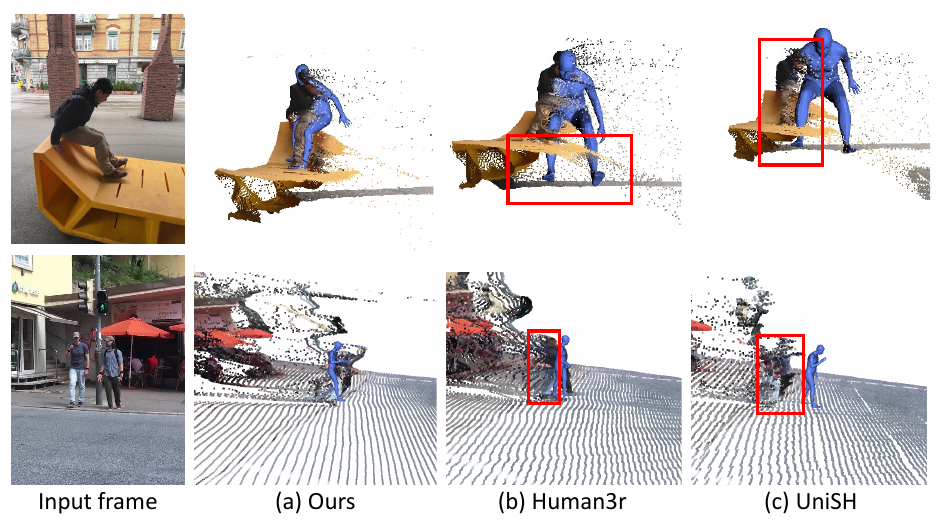}
\caption{
Qualitative comparison of human point map and human body model alignment for a single image. Given a monocular human-centric video (top), we compare our method with Human3r and UniSH. Our method achieves accurate alignment with the human point map. In contrast, Human3r and UniSH exhibit either large-scale misalignment or fail to align with the human point map.
}
  \Description{pipeline}
    \label{fig:single_frame}
\end{figure*}

\paragraph{Human scene alignment.}
We compare our method with recent joint human–scene feed-forward approaches, as shown in Table~\ref{tab:human_scene_metrics}. Our method achieves the best performance on most metrics among approaches that do not explicitly predict metric point maps. We attribute this to our mask-prompting mechanism and end-to-end optimization. In contrast, UniSH freezes both the geometry and HMR branches and relies on a lightweight alignment network to estimate scale and human translation. Our method instead fine-tunes the visual geometry backbone, aligning its depth distribution with the training data and enabling stronger coupling between scene geometry and human parametric priors. As a result, even without additional real-world video datasets, we achieve better human–scene consistency than UniSH. Human3R employs a metric geometry backbone trained on large-scale metric data, which leads to strong performance in HS-CF. However, it injects human tokens only at the decoder stage to infer spatial location, without fully enforcing consistency between human and scene representations. This often results in poor scale alignment, reflected in inferior HS-V scores (see Fig.~\ref{fig:human_scene}, Fig.~\ref{fig:human_point} and Fig.~\ref{fig:single_frame}). In contrast, our ray encoder explicitly conditions human prediction on scene intrinsic information, which is critical for achieving coherent human–scene reconstruction.

\subsection{Ablation Study}
\begin{figure*}[!t]
    \centering
\includegraphics[width=\linewidth]{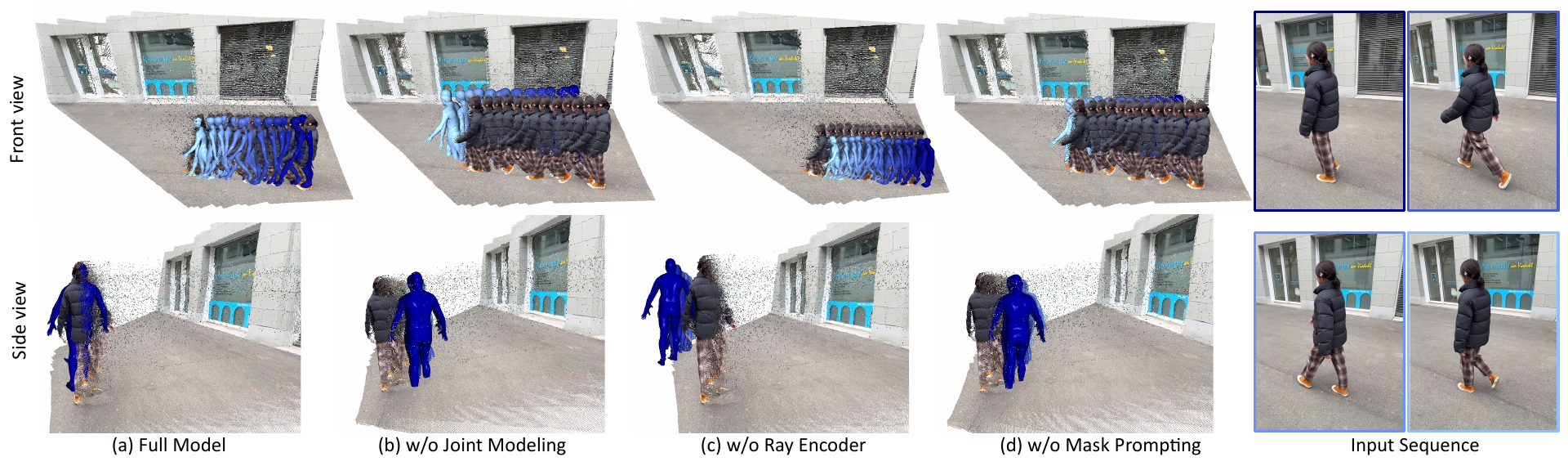}
\caption{Ablation study of our design choices. (a) Full model prediction, where the human is well aligned with the scene point cloud. (b) When scale is estimated solely from scene semantics without the human branch, the model fails to place the human and scene in a consistent global space. (c) Without ray encoder, human branch cannot effectively incorporate scene information or recover consistent camera intrinsics, leading to misalignment. (d) Without mask prompting, scale prediction is affected by irrelevant regions, resulting in suboptimal alignment between the human and the scene. }
  \Description{pipeline}
    \label{fig:ablation}
\end{figure*}

As shown in Tab~\ref{tab:ablation}, we conduct throughly ablation experiments on 3DPW dataset to validate the effectiveness of our proposed human scene reconstruction method. Visualization results of ablation study are shown in Fig.~\ref{fig:ablation}.

\paragraph{Pretraining Alignment}
We observe that removing pretraining alignment significantly degrades performance across all metrics ($HS\text{-}V_5$: 0.708 vs. 0.028, $HS\text{-}CF_5$: 38.51 vs. 9.44). During training, we normalize the predicted point maps at each step (e.g., enforcing a unit mean), which defines the scale convention used by our scale prediction branch. However, the pretrained visual geometry backbone (VGGT), which has not been trained on BEDLAM, produces point maps with a different scale distribution. This mismatch leads to inconsistency between the backbone outputs and the scale branch, making scale recovery difficult. Therefore, finetuning is necessary to align the backbone’s geometry predictions with the normalized distribution used in our training pipeline.

\paragraph{Mask Prompt for visual geometry model.}
Removing mask prompting leads to a clear degradation in performance ($HS\text{-}CF_5$: 14.12 vs. 9.44), indicating that mask prompt guidance is crucial for accurate human–scene alignment.

\paragraph{Ray Encoder}
The ray encoder plays a crucial role in aligning human and scene representations within a shared space. Removing it leads to a substantial degradation in HS-V (0.734). The visual geometry backbone, pretrained on large-scale datasets, is capable of accurately estimating camera intrinsics. The ray encoder leverages this property by injecting intrinsic-aware information into the human branch, thereby ensure the same camera-intrinsic information is shared between scene and human branch to enable better alignment.

\paragraph{Joint modeling of Human and Scene}
We ablate the joint optimization by removing the human branch and estimating scale directly from visual geometry backbone semantics. This results in significant degradation, especially in HS-CF, indicating poor metric-scale alignment. The results show that scene-only cues are insufficient to resolve scale ambiguity, while joint modeling with human-parametric model constraints is crucial for accurate human–scene alignment.

\paragraph{End-to-End Optimization}
Finally, we evaluate end-to-end (E2E) joint tuning of the human and scene branches. This setting achieves the best overall performance, significantly improving $HS\text{-}CF_5$ (from 9.44 to 6.21) and $HS\text{-}CF_{10}$ (from 9.38 to 5.99). This demonstrates that joint optimization further refines metric scale alignment, enabling more consistent reconstruction between human and scene.

\section{Conclusions}
We presented \textbf{SHOW}, a unified feed-forward framework for joint human–scene reconstruction in a shared metric space from monocular video. Unlike prior methods that treat scene reconstruction and human mesh recovery separately, our approach integrates both tasks within a single model to ensure consistent reasoning over geometry, camera pose, and human motion.

At its core, SHOW introduces a mask-as-prompt mechanism that uses human segmentation to guide a pretrained geometry foundation model toward human-relevant 3D structures, reducing reliance on heuristic localization cues. A unified decoder jointly models scene geometry and SMPL-X parameters, producing metrically aligned and globally consistent reconstructions.

Experiments on 3DPW, EMDB, and RICH show that SHOW improves both local human reconstruction accuracy and human–scene alignment within a feedforward model. Overall, SHOW offers a simple yet effective framework for unified 4D perception of humans and their environments.

\section{Limitation and Future Work}
While our method is capable of reconstructing humans and scenes within a unified metric space, a sim-to-real generalization gap still persists. Despite being trained on large-scale synthetic human–scene datasets, the model struggles to generalize to challenging real-world video scenarios. In particular, out-of-distribution cases—such as scenes with extreme scale variations or situations where the human occupies only a very small portion of the image—remain difficult for the model to handle robustly.
\bibliographystyle{ACM-Reference-Format}
\bibliography{bibliography}
\appendix





\begin{figure*}
    \centering
    \includegraphics[width=0.9\linewidth]{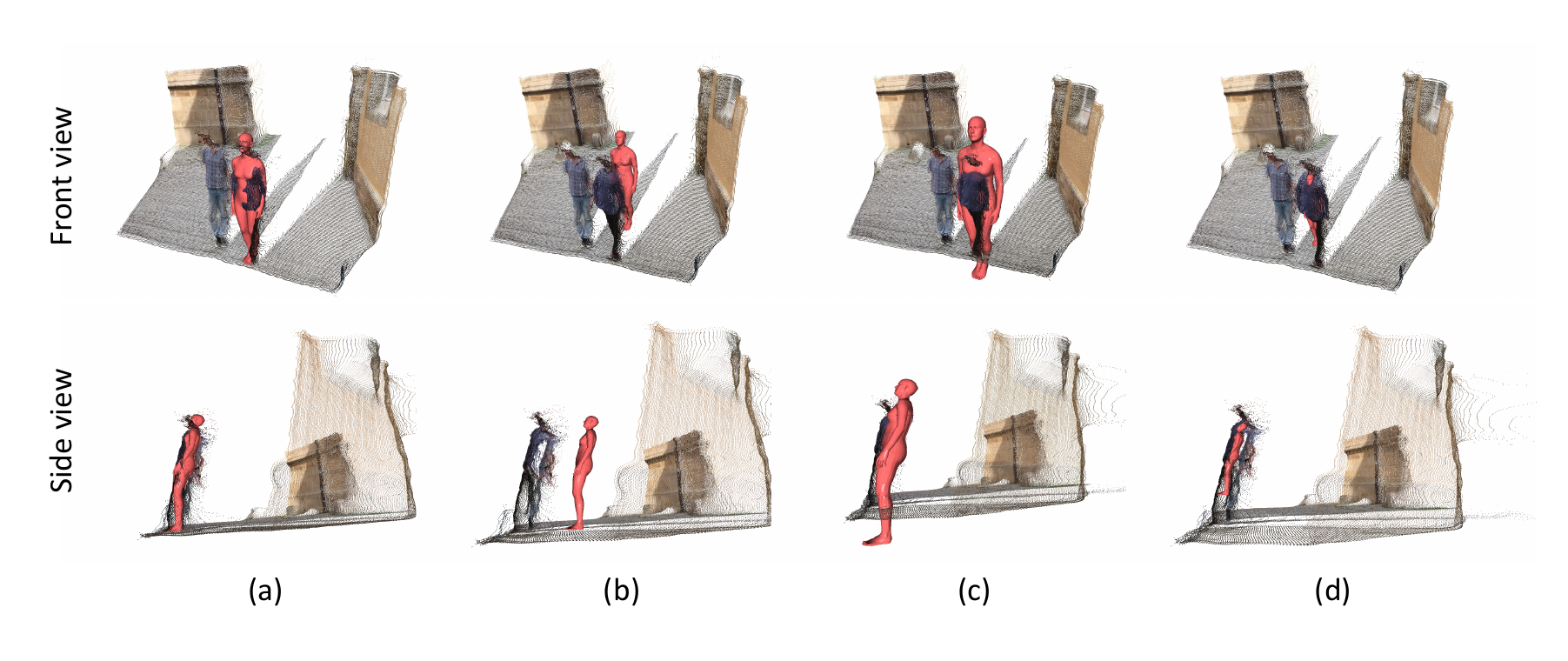}
    \vspace{-20pt}
    \caption{Intuitive visualization of human–scene metrics: Human–Scene Chamfer (HS-CF) and Human–Scene Variance (HS-V). (a) The human is well aligned with the scene, resulting in low HS-CF and HS-V. (b) The human is at an incorrect location but has the correct scale, leading to high HS-CF but low HS-V. (c,d) The human is correctly located but exhibits scale mismatch, resulting in low HS-CF but high HS-V.}
    \Description{metric visualization}
    \label{fig:metric}
\end{figure*}

\section{Evaluation Metric}

Here, we provide a detailed explanation and intuitive visualization of the proposed human–scene alignment evaluation metrics: Human-Scene-Chamfer (HS-CF) and Human-Scene-Variance (HS-V).

Since the reconstructed scene is represented in its most general form as a point cloud, and the input images and corresponding point maps may vary in resolution and include invalid points, we adopt a unified 3D representation using point clouds of size N×3, rather than grid-based representations. For the human component, different parametric body models can be used, such as SMPL, SMPL-X, and MHR \cite{yang2026sam3dbody}. We render the human mesh into the camera frame to obtain a corresponding point map. Both HS-CF and HS-V are then computed between the filtered point clouds, where outliers and invalid points are removed.

This formulation is general and extensible. It can be naturally applied to other parametric representations, including rigid, articulated, and deformable (soft) entities, making it a versatile metric for broader scene co-reconstruction tasks.

As shown in Fig.\ref{fig:metric}, we provide intuitive visualizations of HS-CF and HS-V. As illustrated in the figure, we present four representative cases corresponding to different HS-CF and HS-V combinations. A low HS-CF indicates that the visible surface of the human model is well aligned with the scene point cloud, whereas a high HS-CF suggests poor spatial alignment, with the human surface deviating significantly from the scene geometry. On the other hand, a low HS-V implies that the human model has a reasonable scale and is consistent with the scene point map. In contrast, a high HS-V indicates scale inconsistency, where the human model may appear unrealistically small or large relative to the scene.

\section{SMPL-X human location representation}

Regressing the location of the human in the camera space is much more challenging than most prior work that models humans in a cropped image space. Therefore, we do not regress $\tau$ directly. We regress focal length normalized 2D translation $p_{xy}\in \mathbb{R}^2$ and inverse depth $p_z \in \mathbb{R}$, and then transform them to $\tau$ as follows
\begin{equation} 
    t_{xy} = \frac{p_{xy}}{p_z} \quad
    t_z = \frac{1}{p_z} \times \frac{f}{f_c} \quad
    \tau = [t_{xy}, tz],
    \label{eq:transl}
\end{equation}
where $f$ is the ground truth or estimated focal length of the image, and $f_c$ is the canonical focal length. Predicting the normalized inverse depth follows the recent monocular depth literature~\cite{ranftl2022midas} and is also intuitive since the inverse depth is linearly related to the size of the human in the image. Predicting $p_{xy}$ is equivalent to predicting the 2D location of the human in a normalized image plane.

\section{3D reconstruction loss}
We train the model $f$ end-to-end using a multi-task loss following VGGT:
\begin{equation}\label{eq:training_loss}
\mathcal{L}
=
\mathcal{L}_\text{camera}
+ \mathcal{L}_\text{depth}
+ \mathcal{L}_\text{pmap}
+ \mathcal{L}_\text{dp}
\end{equation}
We describe each loss term in turn.

The camera loss $\mathcal{L}_\text{camera}$ supervises the cameras $\hat{g}$:
$
\mathcal{L}_\text{camera} = \sum_{i=1}^N 
\left \| \hat{g}_i - g_i 
\right \|_\epsilon,
$
comparing the predicted cameras $\hat{g}_i$ with the ground truth $g_i$ using the Huber loss $|\cdot|_\epsilon$.

The depth loss and point map loss follows VGGT loss definition.
Hence, the depth loss is
$
\mathcal{L}_\text{depth}
=
\sum_{i=1}^N
\| \Sigma_i^D \odot (\hat{D}_i - D_i) \| + \| \Sigma_i^D \odot ({\nabla} \hat{D}_i - {\nabla} D_i) \|
- \alpha \log \Sigma_i^D
$,
where $\odot$ is the channel-broadcast element-wise product. 
The point map loss is defined analogously but with the point-map uncertainty $\Sigma_i^P$:
$
\mathcal{L}_\text{pmap}
=
\sum_{i=1}^N
\| \Sigma_i^P \odot (\hat{P}_i - P_i) \| + \| \Sigma_i^P \odot ({\nabla} \hat{P}_i - {\nabla}  P_i) \|
- \alpha \log \Sigma_i^P$.

\end{document}